\documentclass{article}
\usepackage{spconf,amsmath,graphicx}
\usepackage{amssymb}
\usepackage[dvipsnames]{xcolor}
\usepackage[linkcolor=OrangeRed, citecolor=NavyBlue,urlcolor=DarkOrchid, colorlinks=true]{hyperref}
\usepackage[capitalize,noabbrev]{cleveref}
\usepackage{booktabs}
\usepackage{colortbl}
\usepackage[font=small,skip=1pt]{caption} 

\newcommand{\etal}{\textit{et al.\hspace{0.5mm}}}

\title{A Unified Transformer-Based Framework with Pretraining \\ For Whole Body Grasping Motion Generation}
\name{Edward Effendy, Kuan-Wei Tseng, Rei Kawakami}
\address{Institute of Science Tokyo}
\begin{document}
%
\maketitle
\begin{abstract}
We present a novel transformer-based framework for whole-body grasping that addresses both pose generation and motion infilling, enabling realistic and stable object interactions. Our pipeline comprises three stages: \emph{Grasp Pose Generation} for full-body grasp generation, \emph{Temporal Infilling} for smooth motion continuity, and a \emph{LiftUp Transformer} that refines downsampled joints back to high-resolution markers. To overcome the scarcity of hand-object interaction data, we introduce a data-efficient \emph{Generalized Pretraining} stage on large, diverse motion datasets, yielding robust spatio-temporal representations transferable to grasping tasks. Experiments on the GRAB dataset show that our method outperforms state-of-the-art baselines in terms of coherence, stability, and visual realism. The modular design also supports easy adaptation to other human-motion applications. The code can be seen \href{https://github.com/grgward108/PosePretrain}{here}.
\end{abstract}
\section{Introduction}
\label{sec:intro}

Whole-body grasping—coordinating an entire human-like body to stably interact with objects—has broad applications in robotics \cite{8967772}, animation, and virtual reality. Yet many existing methods remain hand-centric, overlooking the vital interplay among hands, arms, torso, and legs necessary for truly stable and natural interactions. By considering the body as a whole, we achieve more realistic balance, better force distribution, and greater adaptability across tasks, paving the way for more immersive and robust human-centric systems.

Whole-body grasping remains a critical yet underexplored area, largely constrained by the scarcity of large-scale motion capture datasets that include both high-fidelity body markers and rich hand-object interactions. While invaluable, existing collections like GRAB~\cite{GRAB:2020} are constrained by their relatively small size, which limits their utility for training data-hungry models such as transformers. However, creating new datasets is prohibitively expensive and time-intensive, requiring specialized equipment, skilled personnel, and significant effort to ensure data quality, further complicating the ability to meet modern deep learning demands.

Efforts to address these challenges have shown promise but leave room for improvement. GOAL~\cite{taheri2022goal} employs a variational autoencoder (VAE) with an autoregressive model for motion infilling, though autoregressive models can struggle with long sequences. SAGA~\cite{wu2022saga} uses dual VAEs and CNNs,  excelling at fine-grained feature extraction, but CNNs tend to prioritize local details over the global dependencies essential for whole-body grasping. These limitations highlight the need for methods that model long-range dependencies while utilizing limited datasets efficiently.

\begin{figure*}[t]
    \centering
    \includegraphics[width=1.0\linewidth]{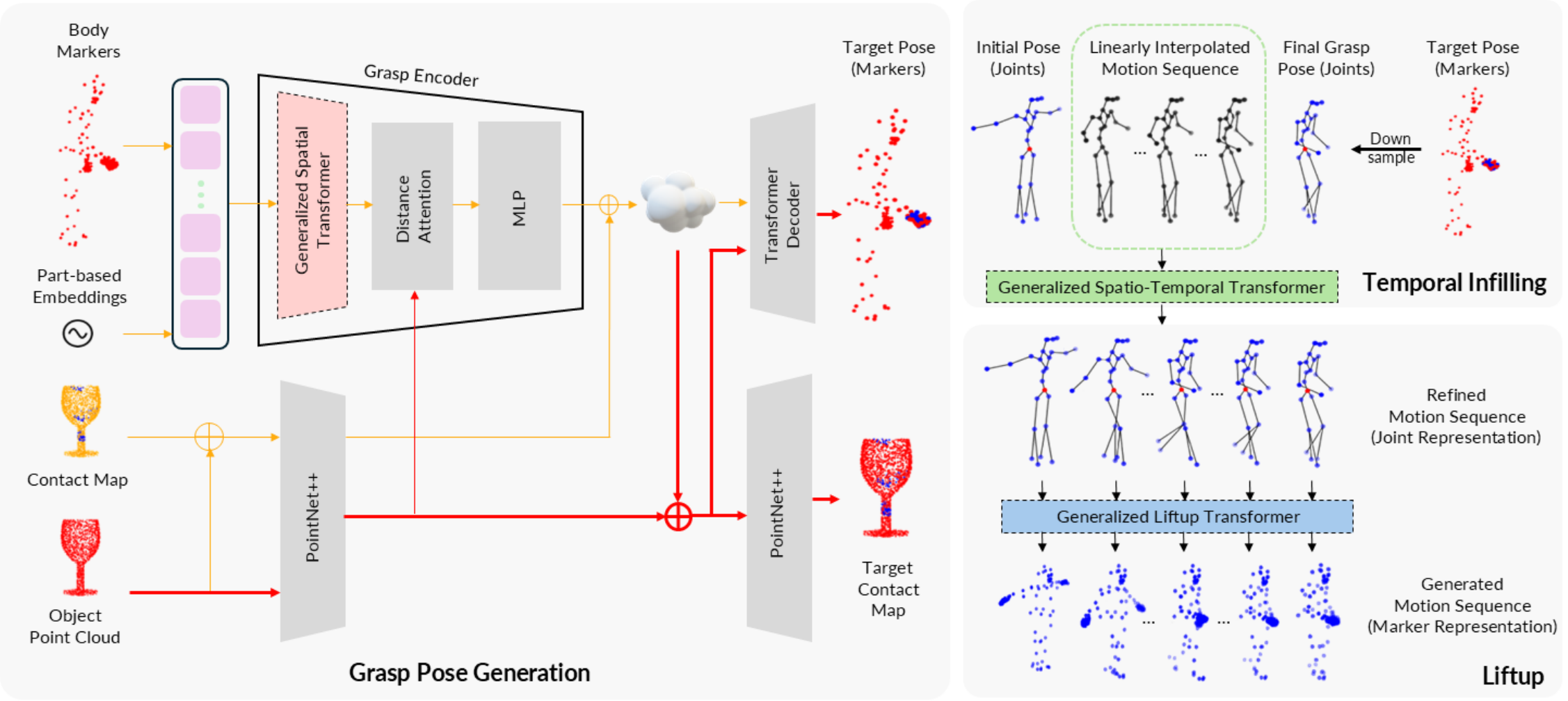}
\caption{\textbf{Three stage pipeline.} Left: Grasp Pose Generation architecture. The model takes object point cloud and body markers to predict grasp points and generate contact maps for both human and object. Lines in red are activated during inference. Top right: Temporal Infilling, where linear interpolation of the first and last pose is refined (red pelvis marker = global context, blue markers = local context). Bottom right: Liftup Transformer, which upsamples joints to the full marker resolution and restores spatial details. The colored boxes are pretrained. } 

    \label{fig:three stage pipeline}
\end{figure*}

To address these limitations, we propose a task-agnostic \textit{Generalized Pretraining} stage, designed to learn general human body dynamics from diverse, non-interactive movement data. By focusing initially on fundamental motion patterns—independent of specific hand-object interactions—the model gains a robust, generalizable representation of human kinematics. We then fine-tune this pretrained transformer on the smaller, specialized grasping dataset, effectively bridging the divide between broad motion understanding and the precise skill of whole-body grasping. 

Our framework is novel in dividing the process into
three transformer-based stages that can be trained and adapted independently. The first stage is \textit{Grasp Pose Generation}, where the model generates a full-body grasping pose based on the given object point cloud. The second is \textit{Temporal Infilling}, which infills  between the starting pose input and the generated pose. These poses are first downsampled to joints and linearly interpolated, then passed through the temporal infilling transformer to produce a motion sequence. Lastly, each frame is individually 
fed to
the \textit{LiftUp Transformer}, which refines the sequence by lifting back the down-sampled joints to the original, high-resolution marker representation. By decoupling these stages, our system remains robust and adaptable, allowing any of the components to be reused or enhanced without a complete overhaul of the pipeline. The pretrained transformers are also reusable for broader applications.

To summarize, the novelty of our work lies in: \textbf{1)} a 
unified 
transformer-based framework for generating realistic and stable whole-body grasping motions, consisting of three stages,
and \textbf{2)} a task-agnostic pretraining strategy to better learn generalized human dynamics. We obtain notable improvements on the GRAB dataset, where we observe fewer motion artifacts and more stable grasps compared to existing methods.

\section{Related Work}
\label{sec:format}

\noindent \textbf{Human Grasp Synthesis.} \hspace{1mm} Grasping methods like GraspTTA \cite{jiang2021graspTTA}, GEARS \cite{zhou2024gears}, DGTR \cite{xu2024dexterous}, and Zhou \etal \cite{Zhou_2024_CVPR} focus on generating realistic hand-object interactions, excelling in fine-grained precision and finger-object alignment. However, they fall short in modeling whole-body dynamics essential for tasks requiring balance, posture adjustments, or locomotion, such as grasping distant objects. 

Whole-body grasping methods address this gap by conditioning on diverse inputs, including text descriptions \cite{diller2023cghoi, ghosh2022imos} or object motion sequences \cite{xu2023interdiff, Zhang2025diffgrasp}. Yet, these approaches are not fully suited to scenarios where the input is limited to a static initial pose and an object, leaving room for improvement in such constrained setups.

\newcommand{\myvspace}{1em}
\noindent \textbf{Motion Infilling.} \hspace{1mm}
Motion infilling, a subset of motion synthesis, generates intermediate motion frames constrained by given past and future frames. Qin \etal \cite{10.1145/3550454.3555454} propose a two-stage transformer pipeline, refining coarse outputs for smooth transitions. Similarly, Oreshkin \etal \cite{oreshkin2022motion} use a transformer encoder-decoder to refine motion deltas from spherical interpolation. Both methods operate at the joint level and rely heavily on multiple context frames for continuity, making them less suitable for sparse input settings. In contrast, Kauffman \etal \cite{MotionInfilling} use convolutional autoencoders to reconstruct masked motion segments from a four-channel image representation, offering an alternative approach. Wang \etal \cite{wang2020synthesizing} introduce Route+PoseNet, an LSTM-based model combining trajectory and pose information for sequential generation. TOHO \cite{Li_2024_WACV} innovates by employing a hypernetwork that generates weights for an Implicit Neural Representation (INR) block, enabling continuous motion infilling without the need for multiple context frames. Our approach is tailored to marker-based representations and constrained by start and end coordinates, providing greater flexibility for grasp-specific tasks.

\noindent\textbf{Transformer Models in Human Motion.} \hspace{1mm}
Transformers \cite{46201} excel at processing sequences in parallel and capturing long-range dependencies, making them highly effective for human motion modeling \cite{
petrovich21actor, Tseng_2025_WACV}. Models like MotionBERT \cite{motionbert2022} and PoseFormer \cite{zheng20213d} combine spatial and temporal attention to lift 2D skeletons to 3D representations, achieving high accuracy in tasks like 3D pose estimation, human mesh recovery, and motion reconstruction. While effective for generating complete motion sequences, these models often struggle with context-aware infilling required for smooth transitions in tasks like grasping. Recent advancements, such as scene-aware transformers \cite{9880007}, incorporate external context into motion predictions, enabling more realistic and adaptable human-object interactions. Inspired by their success in capturing spatial-temporal relationships, we adopt transformers to tackle sparse inputs and generate grasp-specific motions.

\begin{figure*}[t]
\centering
\includegraphics[width=1.0\linewidth]{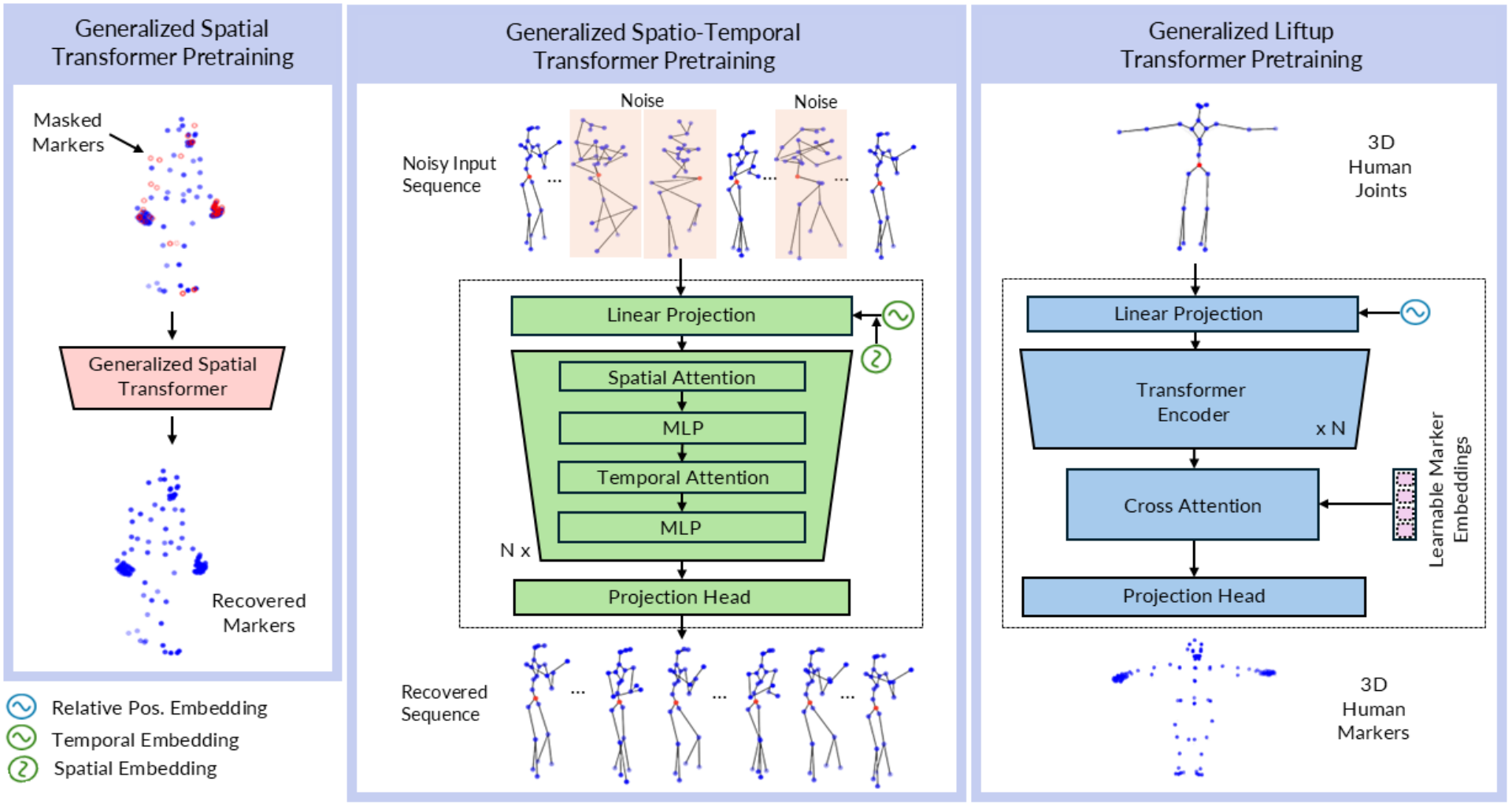}
\caption{\textbf{Generalized Pretraining Stage.} Left: Generalized Spatial Transformer pretraining, recovering recover masked human markers. 
Middle: Generalized Spatio-Temporal Transformer pretraining, recovering noisy human clips. Right: Generalized Liftup Transformer pretraining, learning to map sparse joints to dense markers. }
\vspace{-2mm}
\label{fig:Generalized Pretraining}
\end{figure*}
\vspace{-2mm}
\section{Proposed Method}
\label{sec:pagestyle}

Given an object point cloud and a starting pose as the input, our framework generates whole-body grasping through a three-stage pipeline, each stage leveraging transformer architectures that can be (pre)trained to capture different aspects of human motion. 
Fig.\ref{fig:three stage pipeline} illustrates the three-stage pipeline. First,  \textit{Grasp Pose Generation} predicts a stable, full-body grasping pose from an object point cloud, ensuring accurate hand-object contact. Second, \textit{Temporal Infilling} creates a smooth transition from the initial pose input to the final grasp pose by operating on a downsampled set of $J$ joints. Finally, the \textit{LiftUp Transformer} processes each frame independently, lifting the low-dimensional joint representation back to the full $N$-marker resolution and restoring the spatial details critical for visually realistic and stable motion.

In addition to these three task-specific modules, we perform a \textit{Generalized Pretraining} stage on large-scale motion data, yielding three core transformer variants. Fig. \ref{fig:Generalized Pretraining} illustrates these architectures: (1) a \emph{Generalized Spatial Transformer}, trained to recover masked markers within individual frames, and (2) a \emph{Generalized Spatio-Temporal Transformer}, trained to reconstruct noisy motion sequences across multiple frames, and (3) a \emph{Generalized Liftup Transformer} to learn how to map sparse markers to dense markers. These pretrained weights can be integrated and fine-tuned in the Grasp Pose Generation and Temporal Infilling and Liftup stages, allowing our pipeline to benefit from robust spatio-temporal representations without relying solely on small grasp-specific datasets.

\subsection{Generalized Pretraining}
We pretrain our model on diverse motion datasets, encompassing activities such as walking, dancing, sports, and daily tasks, enabling it to learn generalizable patterns across various action classes.

\begin{figure*}[t]
\centering
\includegraphics[width=0.95\linewidth]{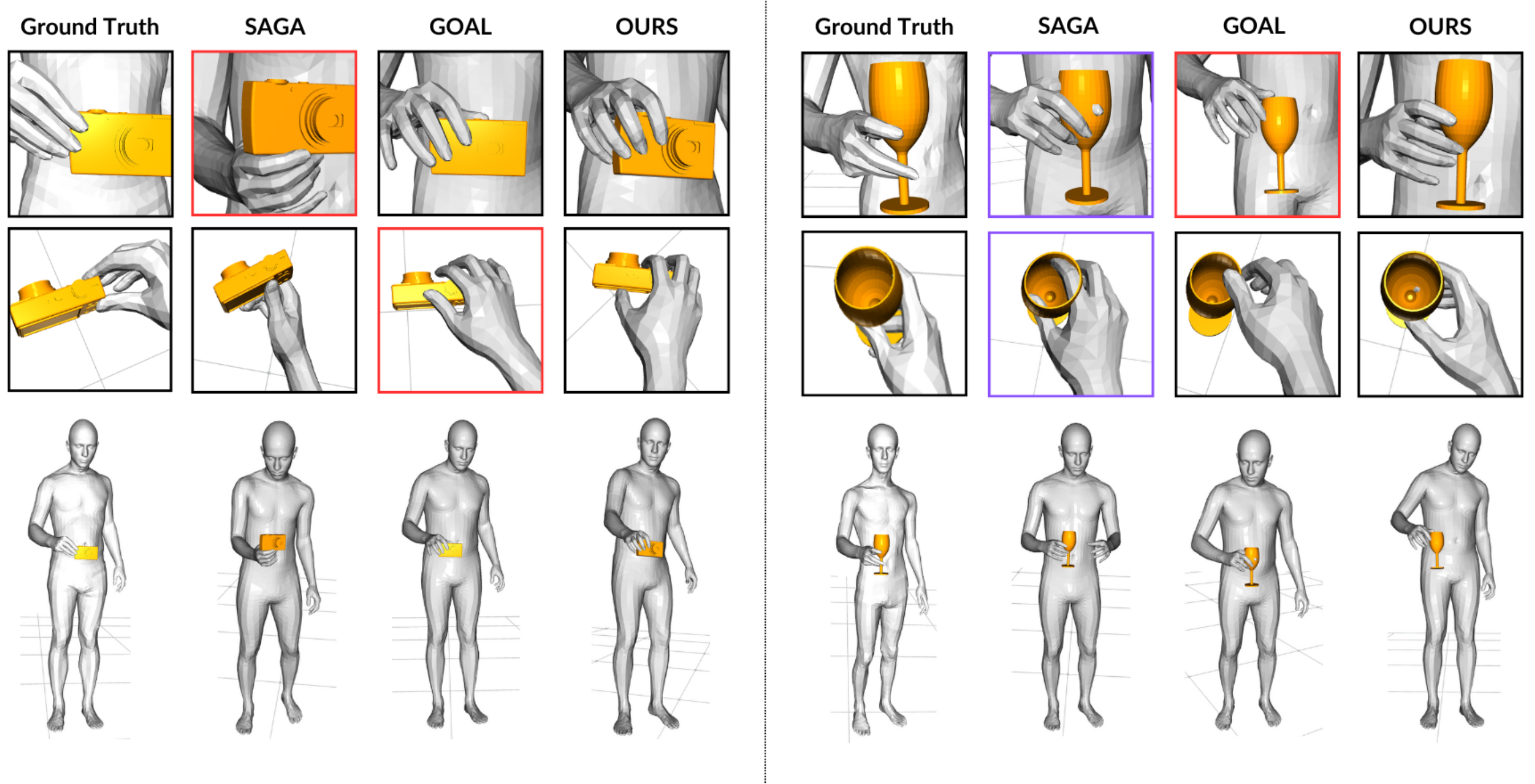}
\caption{\textbf{Qualitative Comparison of Grasp Pose Generation on GRAB \cite{GRAB:2020} dataset}. We compare GOAL \cite{taheri2022goal}, SAGA \cite{wu2022saga}, and the proposed method with the ground truth. Artifacts are marked in different colors, with lack of contact points during grasping marked as red (GOAL \cite{taheri2022goal}, both instances) and hand-object interpenetration (SAGA \cite{wu2022saga}, wineglass) marked as purple. Refer to the \href{https://drive.google.com/drive/folders/1NysaqGKYt8vNhUxP7SNUND8Q_Itg5V94?usp=sharing}{appendix} for more results. }
\label{fig:results}
\end{figure*}

\noindent \textbf{Generalized Spatial Transformer.} The goal here is to capture intra-frame marker relationships by recovering masked markers as shown in  Fig.~\ref{fig:Generalized Pretraining} Left. For each frame, we mask a subset of markers and train the model to reconstruct their positions. This fosters a robust understanding of human body geometry in a purely spatial context. We optimize the loss, defined as:
$
\mathcal{L}_{\text{Spatial}} = \frac{1}{N} \sum_{i=1}^N \| \mathbf{m}_i - \hat{\mathbf{m}}_i \|
\label{eq:spatial_loss}
$
\noindent where \(\mathbf{m}_i\) and \(\hat{\mathbf{m}}_i\) denote the ground truth and reconstructed positions for the \(i\)-th marker, and total markers of \(N\).

\noindent \textbf{Generalized Spatio-Temporal Transformer.}  
The model learns dynamic joint relationships by recovering motion sequences degraded with Gaussian noise, leveraging temporal and spatial attention inspired by MotionBERT~\cite{motionbert2022}, as in  Fig.~\ref{fig:Generalized Pretraining} Middle. The total loss is defined as: $
\mathcal{L}_{\text{Total}} = \alpha \mathcal{L}_{\text{Recon}} + \beta \mathcal{L}_{\text{Vel}} + \gamma \mathcal{L}_{\text{Accel}} + \delta \mathcal{L}_{\text{Pelvis}} + \zeta \mathcal{L}_{\text{Foot}}$
where \(\alpha\), \(\beta\), \(\gamma\), \(\delta\), and \(\zeta\) are hyperparameters. Reconstruction loss \(\mathcal{L}_{\text{Recon}}\) is a weighted MSE between ground truth \(\mathbf{y}_{t, j}\) and predicted positions \(\hat{\mathbf{y}}_{t, j}\) defined as  
{\setlength{\abovedisplayskip}{1pt}
\setlength{\belowdisplayskip}{1pt}
\begin{equation}
\mathcal{L}_{\text{Recon}} = \frac{1}{W} \sum_{t=1}^T \sum_{j=1}^J w_j \|\mathbf{y}_{t, j} - \hat{\mathbf{y}}_{t, j}\|^2,
\label{eq:reconstruction_loss}
\end{equation}}
with \(T\) frames, \(J\) joints, joint weight \(w_j\) and \(W = \sum_{j=1}^J w_j\). Velocity loss \(\mathcal{L}_{\text{Vel}}\) and acceleration loss \(\mathcal{L}_{\text{Accel}}\) extend \(\mathcal{L}_{\text{Recon}}\) to first and second temporal derivatives, ensuring smooth transitions. Pelvis loss $\mathcal{L}_{\text{Pelvis}}$ is also a weighted MSE between the predicted and ground truth pelvis positions. Foot-skating loss \(\mathcal{L}_{\text{Foot}}\) reduces sliding artifacts during foot contact
{\setlength{\abovedisplayskip}{1pt}
\setlength{\belowdisplayskip}{1pt}
\begin{equation}
\mathcal{L}_{\text{Foot}} = \frac{1}{N_f} \sum_{t=1}^{T-1} \sum_{k \in \text{feet}} \|\mathbf{f}_{t+1, k} - \mathbf{f}_{t, k}\|^2,
\label{eq:foot_loss}
\end{equation}}
where \(N_f\) is the number of foot-contact points, and \(\mathbf{f}_{t, k}\) represents the position of foot joint \(k\) at time \(t\).

\noindent \textbf{Generalized LiftUp Transformer.} This model transforms sparse joint positions into a detailed marker-based human representation. It employs learnable marker embeddings to capture the fine-grained spatial relationships between joints and markers. These marker embeddings interact with encoded joint features through a cross-attention mechanism, enabling the model to generate marker-specific representations, as seen in Fig. \ref{fig:Generalized Pretraining} Right. The model uses an \(L1\) loss similar to the Generalized Spatial Transformer.

\vspace{-3mm}
\subsection{Three Stage Pipeline}

\noindent \textbf{Grasp Pose Generation.}  
This module generates whole-body grasping poses from an object point cloud using a Transformer-VAE model as in Fig.~\ref{fig:three stage pipeline} Left. We fine-tune the Generalized Spatial Transformer within the decoder before the VAE phase. We also implement \textit{distance-based attention,} inspired by GOAL~\cite{taheri2022goal}, which is a cross-attention to model body-object relationships, defined as \(I_w(\mathbf{d}) = e^{-w \mathbf{d}}\), where \(w > 0\) is a scalar weight, and $\mathbf{d}$ represents marker-to-object distances, prioritizing markers involved in object interactions.
\noindent Following SAGA~\cite{wu2022saga}, the training objective includes reconstruction loss (combining positional terms and BCE for contact probabilities), KL divergence to regularize the latent space, and consistency loss to align predicted and ground truth marker-object distances.


\vspace{1mm}
\noindent\textbf{Temporal Infilling.}  
This module generates motion between the start pose input and the generated grasping pose. To reduce complexity and prevent overfitting, \(N\) markers are downsampled to \(J\) joints, linearly interpolated over \(F\) frames, and refined using the Temporal-Infilling Transformer to capture long-range dependencies. We fine-tune the Generalized Spatio-Temporal Transformer on the GRAB dataset. The input comprises interpolated joint positions \(\mathbf{M} \in \mathbb{R}^{T \times J \times 3}\) and the global pelvis trajectory \(\mathbf{P} \in \mathbb{R}^{T \times 1 \times 3}\), concatenated as \(\mathbf{X} \in \mathbb{R}^{T \times (J+1) \times 3}\). The loss functions follow those used in Generalized Spatio-Temporal Pretraining.

\vspace{1mm}
\noindent\textbf{LiftUp Transformer.}  
This module processes each frame individually, mapping the sequence of \(J\) joints generated by Temporal Filling to \(M\) markers as in Fig.\,\ref{fig:three stage pipeline} Right Down. The training loss is  
$
\mathcal{L}_{\text{LiftUp}} = \frac{1}{N} \sum_{i=1}^N w_i \|\mathbf{m}_i - \hat{\mathbf{m}}_i\|^2,
$
where \(N\) is the total number of markers, \(\mathbf{m}_i\) and \(\hat{\mathbf{m}}_i\) are ground truth and predicted positions, and \(w_i\) emphasizes hand markers.


\vspace{-3mm}
\section{Experiments}
\vspace{-3mm}
\subsection{Datasets}

\noindent\textbf{AMASS \cite{AMASS:2019}} is a large-scale dataset consolidating motion capture data from 15 sources, covering diverse motions like walking, dancing, and sports, with over 40 hours of data. We use AMASS for our pretraining.

\noindent\textbf{GRAB \cite{GRAB:2020}} is a human-object interaction dataset with whole-body motion sequences, detailed body and hand pose annotations, and contact labels, enabling precise grasp modeling. We use GRAB for our three-stage pipeline, training on the same split as SAGA \cite{wu2022saga}.

\subsection{Grasp Pose Generation}

\noindent\textbf{Quantitative Results.}
We evaluate 150 grasping pose samples using (1) \textit{contact ratio} (percentage of human markers in contact with the object), (2) \textit{interpenetration depth} (object-human intersections), and (3) \textit{diversity} (average pairwise distance, APD). Results in table \ref{tab:performance_grasp} show that our framework achieves higher contact ratios, lower interpenetration depth, and greater diversity compared to baseline methods, especially when using the Generalized Spatial Transformer.

\noindent\textbf{Human Assessment.}
To evaluate pose naturalness,  As shown in Table~\ref{tab:human evaluation}, while our model's scores are slightly lower than the ground truths, it consistently outperforms GOAL and SAGA in perceived naturalness.

\noindent\textbf{Qualitative Results.}
Figure~\ref{fig:results} provides a visual comparison of our results against GOAL~\cite{taheri2022goal}, SAGA~\cite{wu2022saga}, and the Ground Truth on the GRAB dataset~\cite{GRAB:2020}. Our method demonstrates a firmer grip around objects compared to GOAL and reduces interpenetration compared to SAGA.
\vspace{-3mm}
\subsection{Motion Generation Results}

\noindent\textbf{Quantitative Results.}
We evaluated the combined Temporal Infilling and LiftUp stages on the GRAB dataset using SAGA metrics: (1) \textit{3D marker accuracy} (average L2 Distance Error for our method, minimal error between ground truth and 10 random samples for SAGA), (2) \textit{motion smoothness} (power spectrum KL divergence, PSKL-J, for both directions), and (3) \textit{foot skating} (heel proximity and speed thresholds).
Results in Table \ref{tab:performance_motion} show that our generalized spatio-temporal transformer outperforms SAGA across metrics. Models trained solely on GRAB data perform worse, highlighting the importance of pretraining for robust results.

\noindent\textbf{Qualitative Results.}
Video comparisons of motion generation results between the SAGA method \cite{wu2022saga} and ours are included in the appendix. \href{https://drive.google.com/drive/folders/1NysaqGKYt8vNhUxP7SNUND8Q_Itg5V94?usp=sharing}{Click here to view the videos}.


\begin{table}[t]
\centering
\caption{Performance metrics comparison. CR: contact ratio (\%), ID: interpenetration depth(cm). We compare the APD of the sampled markers, and the CR and ID of the grasping hand vertices. Numbers in block indicate best results.}
\label{tab:performance_grasp}
\begin{tabular}{l c c c}
\toprule
Method & APD ($\uparrow$) & CR ($\uparrow$) & ID ($\downarrow$) \\
\midrule
GOAL & 1.55 & 2.6 & 14.81  \\
SAGA & 4.19 & 5.2 & 3.94 \\
\textbf{Ours} (from scratch) & 4.86 & 5.2 & 3.32 \\
\textbf{Ours} (with ST) & \textbf{7.19} & \textbf{5.4} & \textbf{2.90} \\
\bottomrule
\end{tabular}
\end{table}

\begin{table}[tbp!]
\centering
\definecolor{yellow}{rgb}{1,1, 0.6}
\definecolor{orange}{rgb}{1, 0.8, 0.6}
\definecolor{red}{rgb}{1, 0.6, 0.6}
\caption{Human evaluation scores for motion realism across different objects, comparing GOAL, SAGA, ours, and ground truth}
\label{tab:human evaluation}
\begin{tabular}{l c c c | c} 
\toprule
\textbf{Objects}  & \textbf{GOAL \cite{taheri2022goal}} & \textbf{SAGA} \cite{wu2022saga} & \textbf{Ours} & \textbf{GT} \\
\midrule
Binoculars &  1.81 & 3.61  &  \textbf{4.04}  &   4.5 \\
Camera     &  2.58 & 2.81  &   \textbf{3.45}  &   3.88 \\
Toothpaste &  2.17 & 3.63  &   \textbf{4.37}  &   4.83 \\
Mug        &  1.59 & 1.97  &   \textbf{2.56}  &   3.59 \\
Wineglass  &  2.33 & 2.13  &   \textbf{3.91}  &   4.29 \\
\bottomrule
\end{tabular}
\end{table}

\begin{table}[tbp!]
\centering
\definecolor{yellow}{rgb}{1,1, 0.6}
\definecolor{orange}{rgb}{1, 0.8, 0.6}
\definecolor{red}{rgb}{1, 0.6, 0.6}
\setlength{\tabcolsep}{4pt} 
\caption{Performance metrics comparison. Bold and underline indicates best and second-best results respectively.}
\label{tab:performance_motion}
\begin{tabular}{l c c c c c}

\toprule
\textbf{Method} & \textbf{ADE ($\downarrow$)} & \textbf{Skat.($\downarrow$)} & \multicolumn{2}{c}{\textbf{PSKL-J} ($\downarrow$)} \\
\cmidrule(lr){4-5}
 & & & \textbf{P→GT} & \textbf{GT→P} \\
\midrule
Lin. Interp. & 0.151 & 0.442 & 0.00 & 15.28 \\
SAGA \cite{wu2022saga} &\underline{0.121} & \underline{0.267} & 1.081 & 0.978 \\
Ours (Scratch) & 0.165 & 0.462 & \underline{0.562} & \textbf{0.641} \\
\textbf{Ours (Pretrain)} & \textbf{0.094} & \textbf{0.124} & \textbf{0.560} & \underline{0.842} \\
\bottomrule
\end{tabular}

\end{table}

\noindent\textbf{Ablation Study.}
We replaced SAGA's explicit foot contact label prediction with a foot-skating loss for contact consistency. Ablation studies (Table \ref{tab:ablation}) revealed that using all loss functions yields the lowest ADE and foot-skating metrics. Although motion smoothness decreases slightly, the visual impact is negligible, making the trade-off worthwhile for improved performance
\vspace{-3mm}
\begin{table}[tbp!]
\centering
\definecolor{yellow}{rgb}{1,1, 0.6}
\definecolor{orange}{rgb}{1, 0.8, 0.6}
\definecolor{red}{rgb}{1, 0.6, 0.6}
\setlength{\tabcolsep}{4pt} 
\caption{Ablation study results showing the impact of individual loss components on evaluation metrics. Removing acceleration and foot-skating losses significantly degrades performance.}
\label{tab:ablation}
\begin{tabular}{l c c c c c}
\toprule
\textbf{Method} & \textbf{ADE ($\downarrow$)} & \textbf{Skat.($\downarrow$)} & \multicolumn{2}{c}{\textbf{PSKL-J} ($\downarrow$)} \\
\cmidrule(lr){4-5}
 & & & \textbf{P→GT} & \textbf{GT→P} \\
\midrule
All loss func. & \textbf{0.094} & \textbf{0.124} & \underline{0.560} & 0.842 \\
-(\( \mathcal{L}_{\text{Acc}} + \mathcal{L}_{\text{Foot}} \)) & \underline{0.108} & 0.300 & 0.603 & \underline{0.760} \\
-\( \mathcal{L}_{\text{Acc}} \) & 0.110 & 0.224 & 0.570 & 0.829 \\
-\( \mathcal{L}_{\text{Foot}} \) & 0.105 & \underline{0.342} & \textbf{0.430} & \textbf{0.561} \\
\bottomrule
\end{tabular}
\end{table}
\vspace{-1mm}


\section{Conclusion and discussion}
\vspace{-3mm}
This paper presents a transformer-based framework for whole-body grasping, integrating generalized spatial, spatio-temporal, and LiftUp Transformers to generate realistic, stable, and smooth human-object interactions. By leveraging generalized pretraining on diverse motion datasets, our approach effectively addresses the challenges posed by limited grasp-specific data, achieving improved performance on the GRAB dataset. The modular design of our framework ensures adaptability to various applications. Future work could focus on generating longer video sequences to capture extended interactions and incorporating scenarios involving multiple objects or humans, further enhancing the versatility and realism of human-machine collaboration.

\vspace{-3mm}
\section{acknowledgements}
\vspace{-2mm}
This work was supported by JSPS KAKENHI grant number JP23H00490. This study was carried out using the TSUBAME4.0 supercomputer at Institute of Science Tokyo.

\bibliographystyle{IEEEbib}
\bibliography{strings,refs}

\end{document}